\documentclass[11pt]{article}
\usepackage{indentfirst}   
\usepackage{graphicx}
\usepackage{amsmath, amssymb}
\usepackage{authblk}
\usepackage{geometry}
\usepackage{datetime}
\newdateformat{mydate}{\THEDAY~\monthname[\THEMONTH]~\THEYEAR}

\title{Domain Generalization with Quantum Enhancement for Medical Image Classification: A Lightweight Approach for Cross-Center Deployment}

\author[1]{Jingsong Xia\thanks{Corresponding author. Email: xiajingsong2@gmail.com}}
\author[1]{Siqi Wang\thanks{Email: wsq03925@163.com}}

\affil[1]{The Second Clinical College, Nanjing Medical University, Nanjing, China}

\date{January 2026}

\begin{document}

\maketitle

\begin{abstract}
Medical image artificial intelligence models often demonstrate excellent performance when trained and evaluated on data from a single center or imaging device. However, their performance frequently degrades substantially during real-world clinical deployment across hospitals, scanners, or imaging protocols due to domain shift, severely limiting clinical scalability and generalizability. To address the challenge of insufficient cross-center generalization in medical imaging, we propose a lightweight domain generalization and quantum-enhanced collaborative learning framework that achieves robust generalization to unseen target domains without relying on real multi-center labeled data.

Specifically, we construct a domain-invariant encoder based on MobileNetV2 and enable cross-domain robust learning through three key components: (1) multi-domain imaging shift simulation, where brightness, contrast, sharpening, and noise perturbations are applied to emulate heterogeneous imaging characteristics across hospitals; (2) domain-adversarial training with a gradient reversal mechanism to explicitly suppress domain-discriminative information in feature representations; and (3) a lightweight quantum feature enhancement layer that employs parameterized quantum circuits to perform nonlinear feature mapping and entanglement modeling on deep features. In addition, a test-time adaptation (TTA) strategy is incorporated during inference to further mitigate distribution shifts in unseen domains.

Experiments conducted on simulated multi-center medical image datasets demonstrate that the proposed method significantly outperforms baseline models without domain generalization or quantum enhancement in previously unseen domains. The variance of domain-specific accuracy is markedly reduced, while TTA further improves the area under the ROC curve (AUC) and sensitivity, validating the potential clinical value of the proposed framework for real-world cross-center deployment. This study represents an early attempt to integrate quantum-enhanced feature mapping into medical image domain generalization, achieving improved robustness under constrained computational resources and providing a feasible paradigm for hybrid quantum--classical intelligent medical imaging systems.
\end{abstract}

\section{Introduction}

Medical image artificial intelligence has long faced persistent and formidable challenges in cross-center generalization. In recent years, deep learning–based models have achieved remarkable progress in medical image classification~\cite{1}, segmentation~\cite{2}, and lesion detection~\cite{3}, demonstrating performance comparable to or even surpassing that of human experts on multiple single-center datasets. However, extensive empirical evidence indicates that once these models are deployed in external medical centers, their performance often degrades substantially, and in some cases, fails entirely. The fundamental cause of this phenomenon lies in the ubiquitous presence of domain shift in real-world clinical environments, where systematic statistical discrepancies exist between training data and data encountered during deployment.

Such discrepancies may arise from multiple sources, including variations in imaging equipment vendors and hardware configurations, inconsistencies in imaging protocols and acquisition parameters, differences in image reconstruction algorithms and post-processing pipelines, as well as shifts in patient demographics and disease prevalence. In realistic clinical scenarios, models are frequently required to operate on data originating from previously unseen hospitals or imaging devices, whose data distributions are entirely unknown during training. As a result, the core assumption underlying conventional deep learning—namely that training and test data are independently and identically distributed—no longer holds, severely limiting model reliability in real-world deployment.

To mitigate cross-center generalization issues~\cite{4}, a variety of technical strategies have been proposed, yet each exhibits notable limitations. One line of research relies on joint training with real multi-center datasets, aiming to enhance model robustness by expanding the training distribution~\cite{5}. In practice, however, this approach is severely constrained by data privacy regulations, ethical approval requirements, and the high cost of cross-institutional data sharing. Another class of methods is based on domain adaptation~\cite{6}, in which models are retrained or fine-tuned using data from the target domain. These methods typically assume that target-domain data are accessible during training, an assumption that is often unrealistic prior to actual clinical deployment. In addition, preprocessing techniques such as style transfer~\cite{7} or intensity normalization~\cite{8} attempt to reduce inter-center variability at the image level, yet the complexity and nonlinearity of real clinical imaging variations make it difficult for such methods to comprehensively account for all potential distributional changes. More critically, most existing approaches primarily focus on data-level or training-strategy-level adjustments, while comparatively little attention has been devoted to examining whether deep feature representations themselves possess sufficient structural richness to support genuinely domain-invariant learning.

Against this backdrop, quantum computing~\cite{9} offers a potentially transformative paradigm for medical image feature modeling. In recent years, quantum machine learning has been recognized for its ability to leverage high-dimensional Hilbert space embeddings, quantum entanglement structures, and non-classical higher-order nonlinear transformations, thereby extending the representational capacity of conventional classical neural networks. For medical imaging tasks characterized by high complexity and pronounced distributional heterogeneity, the incorporation of quantum-enhanced feature mappings holds promise for improving the capture of latent structural information and, consequently, enhancing generalization performance in unseen domains. Although current general-purpose quantum hardware remains limited by scale and noise, the development of lightweight, simulatable quantum modules~\cite{10,11,12,13} that can be seamlessly embedded within classical deep networks as feature enhancement components represents a practically feasible pathway at this stage.

Motivated by these considerations, this study proposes a lightweight cross-center medical image classification framework tailored for real-world clinical deployment. By integrating quantum-enhanced modules into a classical deep learning architecture, the proposed framework enables efficient modeling of domain-invariant discriminative features and incorporates interpretable multi-domain imaging perturbation mechanisms to facilitate unsupervised domain generalization without requiring access to target-domain data. Furthermore, the framework is explicitly designed with computational resource constraints in mind, allowing end-to-end training and inference to be performed on laptop-level hardware. To further improve robustness under previously unseen imaging distributions, a test-time adaptation strategy is additionally introduced. Together, these components provide a practical and efficient solution for deploying hybrid quantum–classical models in cross-center medical imaging applications.

\section{Methods}

\subsection{Overall Framework}

One of the core challenges in multi-center medical image analysis lies in the systematic degradation of model generalization caused by imaging distribution shifts. In real-world clinical environments, even for the same disease category, substantial variations in image appearance are commonly observed across different medical centers due to differences in scanning devices, imaging protocols, post-processing algorithms, and operational practices. Importantly, such variations do not reflect underlying pathological differences, yet they are easily misinterpreted by deep learning models as discriminative cues, leading to severe performance degradation when models are deployed across centers.

To address this challenge, we propose a lightweight quantum--classical hybrid learning framework that is explicitly designed to be robust against multi-center imaging distribution shifts. From a system-level perspective, the framework decouples \emph{imaging style modeling}, \emph{disease semantic representation}, \emph{nonlinear feature enhancement}, and \emph{cross-domain generalization constraints}, and integrates them through end-to-end joint optimization. The overall architecture consists of four complementary components: (1) a multi-domain imaging shift simulation module; (2) a domain-invariant feature encoder; (3) a lightweight quantum feature enhancement layer; and (4) a domain-adversarial discrimination mechanism with test-time adaptation.

Formally, given an input medical image $x \in \mathbb{R}^{H \times W \times C}$, the model first applies explicit virtual imaging perturbations via the multi-domain augmentation module to generate domain-tagged samples $x^{(d)}$. The key motivation of this design is to actively expose the model to diverse imaging conditions, rather than passively relying on a single-center data distribution. The perturbed input is then processed by a parameterized feature encoder $f_{\theta_f}(\cdot)$ to extract a shared latent representation:
\begin{equation}
h = f_{\theta_f}\left(x^{(d)}\right), \quad h \in \mathbb{R}^{256}.
\end{equation}

The latent feature $h$ is not merely optimized for disease classification; instead, it is explicitly constrained to be insensitive to the imaging domain $d$ while remaining highly discriminative with respect to the disease label $y$. In other words, the objective of the encoder is not to maximize training accuracy alone, but to learn a stable disease representation that is invariant across device- and protocol-induced variations. To this end, a disease classification head $g_{\theta_y}(\cdot)$ and a domain discriminator $q_{\theta_d}(\cdot)$ are jointly introduced, leading to the following adversarial min--max optimization objective:
\begin{equation}
\min_{\theta_f, \theta_y} \; \max_{\theta_d} \; \mathcal{L}_{\text{cls}}(y, \hat{y}) - \lambda \, \mathcal{L}_{\text{dom}}(d, \hat{d}),
\end{equation}

where $\mathcal{L}_{\text{cls}}$ drives the learning of disease-relevant discriminative features, while $\mathcal{L}_{\text{dom}}$ penalizes the presence of domain-identifiable imaging style information in the latent representation. The hyperparameter $\lambda$ controls the trade-off between disease discriminability and domain invariance. Through this adversarial objective, the framework explicitly disentangles disease semantics from imaging style at the optimization level, providing a stable and well-controlled feature distribution for subsequent quantum feature enhancement.

\subsection{Multi-Domain Imaging Shift Simulation}

In real clinical data, imaging distribution shifts rarely occur as discrete factors; instead, they manifest as continuous and multi-dimensional variations. For example, brightness changes may arise from exposure differences, contrast variations are often related to reconstruction algorithms, and noise levels are closely associated with hardware aging or low-dose acquisition. When trained on data from a single or limited number of centers, deep models are prone to erroneously encoding such non-pathological factors as discriminative features.

To systematically emulate these complex imaging shifts, we construct three virtual imaging domains during training, corresponding to high-end devices, mainstream devices, and legacy equipment. To prevent the model from trivially associating domain labels with discrete identifiers, continuous parameter perturbations are introduced within each virtual domain. Specifically, domain-dependent transformation operators are applied to the original input image $x$:
\begin{equation}
x^{(d)} = T_d(x; \boldsymbol{\phi}_d),
\end{equation}

where the parameter vector $\boldsymbol{\phi}_d$ is defined as
\begin{equation}
\boldsymbol{\phi}_d = (\beta_d, \kappa_d, \nu_d),
\end{equation}

with $\beta_d$, $\kappa_d$, and $\nu_d$ controlling brightness scaling, contrast modulation, and noise intensity, respectively. These parameters are not fixed but sampled from uniform distributions:
\begin{equation}
\boldsymbol{\phi}_d \sim \mathcal{U}(\boldsymbol{\phi}_d^{\min}, \boldsymbol{\phi}_d^{\max}).
\end{equation}

This design ensures substantial intra-domain diversity in imaging style, forcing the model to abandon reliance on low-level texture or intensity statistics for domain discrimination. By continuously introducing such perturbations during training, the model is systematically guided to focus on disease-related structural patterns that remain stable across a continuous imaging shift space, thereby substantially enhancing cross-center generalization.

\subsection{Domain-Invariant Feature Encoder}

In designing the feature encoder, a balance is struck between representational capacity and clinical deployability. MobileNetV2 is adopted as the backbone network due to its use of depthwise separable convolutions, which significantly reduce parameter count and computational complexity. This choice enables stable model operation under resource-constrained environments, aligning with practical clinical deployment requirements.

The encoder performs hierarchical feature extraction on the perturbed input $x^{(d)}$, followed by global average pooling to produce a compact global representation:
\begin{equation}
h = \text{GAP}\left(f_{\text{MB}}\left(x^{(d)}\right)\right).
\end{equation}

To prevent feature magnitude inflation or distributional instability during adversarial training, an additional feature norm regularization term is introduced. This constraint explicitly limits the energy of the latent representation, ensuring that the quantum rotation angles induced by $h$ remain within a smooth regime and that gradients remain stable during classical--quantum hybrid backpropagation:
\begin{equation}
\mathcal{L}_{\text{feat}} = \mathbb{E}\left[\|h\|_2^2\right].
\end{equation}

\subsection{Lightweight Quantum Feature Enhancement}

\subsubsection{Quantum Encoding and Variational Circuit Construction}

Conventional deep networks operating in low-dimensional compact feature spaces are often constrained by linear or weakly nonlinear mappings. To enhance feature expressiveness without significantly increasing model complexity, we introduce a lightweight quantum feature enhancement layer that maps classical features into quantum state space for nonlinear reparameterization. Specifically, quantum rotation angles are generated through linear projection followed by hyperbolic compression:
\begin{equation}
\boldsymbol{\theta} = \pi \cdot \tanh\left(W_q h\right),
\end{equation}

where $W_q \in \mathbb{R}^{n \times 256}$ defines the mapping from classical features to $n$ qubits. The hyperbolic tangent function effectively suppresses extreme input values, naturally constraining the rotation angles within a physically interpretable range.

Based on these parameters, a variational quantum circuit composed of single-qubit rotation gates and a ring-shaped entanglement structure is constructed, yielding the output quantum state $|\psi(\boldsymbol{\theta})\rangle$. This circuit design balances expressive capacity with circuit depth, thereby mitigating sensitivity to noise and simulation overhead.

\subsubsection{Quantum--Classical Residual Fusion}

To effectively integrate quantum information back into the classical network, an expectation-based measurement strategy is employed. Specifically, the Pauli-$Z$ observable is measured on each qubit:
\begin{equation}
z_i = \langle \psi(\boldsymbol{\theta}) | \sigma_z^{(i)} | \psi(\boldsymbol{\theta}) \rangle.
\end{equation}

These measurements yield deterministic real-valued outputs that are seamlessly compatible with classical neural networks. All measurement results are aggregated into a vector $z \in \mathbb{R}^n$, serving as the quantum feature representation. The quantum features are then projected back into the classical feature space and fused with the original latent representation via a residual connection:
\begin{equation}
h' = h + \alpha \cdot W_r z,
\end{equation}

where $\alpha$ controls the contribution of the quantum features. This residual fusion strategy ensures that the quantum module acts as a mild enhancement to the original feature distribution rather than forcibly reshaping the discriminative structure, thereby improving feature diversity while preserving overall model stability and interpretability.

\subsection{Domain-Adversarial Training and Test-Time Adaptation}

To further suppress residual domain-specific information in the latent representation, a domain-adversarial training mechanism is employed. A Gradient Reversal Layer (GRL) is inserted between the feature encoder and the domain discriminator, reversing the sign of gradients propagated from the domain branch during backpropagation. This encourages the encoder to produce features that are indistinguishable from the perspective of domain discrimination.

The strength of the domain-adversarial constraint is dynamically scheduled rather than fixed. A smooth sigmoid-shaped schedule is adopted:
\begin{equation}
\lambda(p) = \frac{2}{1 + \exp(-10p)} - 1,
\end{equation}

where $p \in [0,1]$ denotes the normalized training progress. This strategy allows the model to focus on learning disease-discriminative features during early training stages, while gradually enforcing domain invariance as training progresses, thereby avoiding premature feature collapse or underfitting.

During inference, since target-domain data are typically unlabeled, a test-time adaptation (TTA) mechanism is further introduced. Only the running statistics of Batch Normalization layers are updated according to:
\begin{equation}
\mu_t = (1 - \eta)\mu_{t-1} + \eta \cdot \mu_{\text{batch}},
\end{equation}

without introducing any additional supervision or retraining. This lightweight adaptation enables self-calibration to target-domain distribution shifts and further enhances robustness and generalization in real-world multi-center clinical settings.

\section{Results}

\subsection{Classification Performance Evaluation}

We comprehensively evaluated the classification performance of the proposed domain-generalized quantum-enhanced model (DG-Quantum) against four baseline models—ResNet18, MobileNetV2, EfficientNet-B0, and SimpleCNN—on the simulated multi-center medical image dataset. Figure~1 presents the confusion matrices for all models, illustrating their predictive performance on positive (Pos, tumor-positive) and negative (Neg, normal) samples.

In each confusion matrix, true labels are displayed along the rows and predicted labels along the columns, reporting the counts and corresponding percentages of true negatives (TN), false positives (FP), false negatives (FN), and true positives (TP). Such visualization is particularly important in medical image classification tasks within biomedical engineering, as it not only reflects overall accuracy but also highlights clinically relevant diagnostic biases. Specifically, false negatives may lead to missed diagnoses with severe clinical consequences, whereas false positives may result in unnecessary follow-up examinations or overtreatment.

The DG-Quantum model demonstrates highly consistent and robust classification behavior. On the test set, it achieves 57 true negatives (95.0\% of negative samples) and only 3 false positives (5.0\%), alongside 59 true positives (98.3\%) and merely 1 false negative (1.7\%), yielding an overall accuracy of 0.9667 (96.67\%). This performance indicates a strong capability to capture domain-invariant discriminative features, effectively mitigating cross-center domain shifts induced by heterogeneous imaging conditions. In the context of medical imaging AI, the high true positive rate is particularly significant, as it implies reliable identification of the vast majority of tumor-positive cases, thereby reducing the risk of missed diagnoses and supporting timely clinical intervention. Simultaneously, the low false positive rate preserves diagnostic specificity, minimizing unnecessary invasive procedures and optimizing healthcare resource utilization.

By comparison, the ResNet18 model achieves 56 true negatives (93.3\%), 4 false positives (6.7\%), 56 true positives (93.3\%), and 4 false negatives (6.7\%), corresponding to an accuracy of 0.9333 (93.33\%). Although its performance remains competitive, the slightly elevated false negative and false positive rates suggest reduced stability under complex domain shifts. This behavior may be attributed to the deeper network architecture, which is more prone to overfitting domain-specific characteristics of the source data.

Further analysis of MobileNetV2 and EfficientNet-B0—both lightweight models with potential advantages in resource-constrained clinical environments—reveals more pronounced limitations. MobileNetV2 yields 56 true negatives (93.3\%), 4 false positives (6.7\%), 49 true positives (81.7\%), and 11 false negatives (18.3\%), with an overall accuracy of 0.8750 (87.50\%). The elevated false negative rate represents a critical drawback in medical image classification, as it may lead to tumor cases being incorrectly classified as normal, delaying diagnosis and treatment. Similarly, EfficientNet-B0 records 52 true negatives (86.7\%), 8 false positives (13.3\%), 53 true positives (88.3\%), and 7 false negatives (11.7\%), also achieving an accuracy of 0.8750 (87.50\%). Its higher false positive rate indicates insufficient specificity for negative samples, suggesting that although its efficient convolutional design improves parameter efficiency, it lacks the nonlinear domain-invariant representational capacity provided by quantum-enhanced feature mapping.

The SimpleCNN baseline exhibits the poorest performance among all models. Its confusion matrix shows 60 true negatives (100.0\%), 0 false positives (0.0\%), only 8 true positives (13.3\%), and 52 false negatives (86.7\%), resulting in an accuracy of 0.5667 (56.67\%). The extremely high false negative rate indicates that this shallow convolutional network fails to generalize across multiple imaging domains and is largely incapable of distinguishing positive cases. Such behavior is unacceptable in biomedical engineering practice, as it would substantially increase clinical misdiagnosis rates.

Overall, the confusion matrix analysis highlights the superiority of the DG-Quantum model in achieving a favorable balance between sensitivity and specificity. By integrating quantum circuit–based feature enhancement with domain-adversarial learning, the proposed approach not only outperforms classical baselines numerically but also provides more reliable and clinically interpretable diagnostic behavior. In line with robustness and generalization requirements emphasized by leading medical imaging AI standards, such as MICCAI and RSNA guidelines, these results support DG-Quantum as a promising framework for cross-center deployment, particularly in healthcare systems where imaging device heterogeneity is prevalent.

\begin{figure}[htbp]
    \centering
    \includegraphics[width=0.85\linewidth]{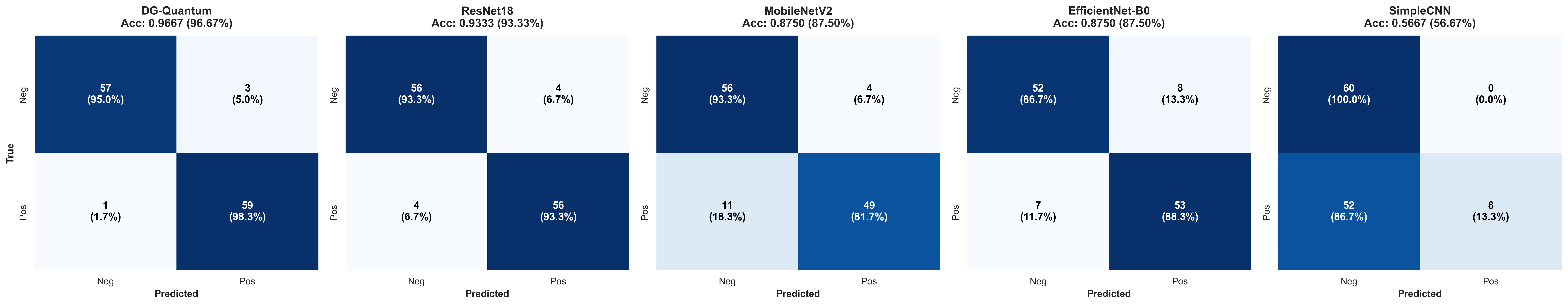}
    \caption{Confusion matrices of classification results across different models.}
    \label{fig:performance_comparison}
\end{figure}

\subsection{Performance Metric Comparison and Confidence Interval Analysis}

To comprehensively quantify model performance and assess statistical robustness, we employed a Bootstrap resampling strategy with $n=500$ iterations to estimate the mean values and corresponding 95\% confidence intervals (CIs) for multiple evaluation metrics, including Accuracy, Area Under the ROC Curve (AUC), F1 score, Sensitivity, and Specificity. The results were visualized in Figures~2, 3, and 4, while detailed numerical values are summarized in Table~\ref{tab:performance_comparison}.

Figure~2 presents point estimates of each metric together with their 95\% CI error bars. Figure~3 provides a multidimensional comparison of Precision, Recall, Sensitivity, Specificity, Accuracy, F1 score, and AUC, where the shaded regions denote the corresponding confidence intervals. Furthermore, Figure~4 illustrates the Bootstrap distributions using the interquartile range (IQR), median, and outliers, enabling a direct assessment of model robustness under repeated resampling.

\begin{table}[htbp]
\centering
\caption{Quantitative comparison of model performance across multiple evaluation metrics.}
\label{tab:performance_comparison}
\begin{tabular}{lccccccc}
\hline
Model & Accuracy & AUC & F1 & Precision & Recall & Sensitivity & Specificity \\
\hline
DG-Quantum        & 0.9667 & 0.9936 & 0.9667 & 0.9672 & 0.9667 & 0.9833 & 0.9500 \\
ResNet18          & 0.9333 & 0.9811 & 0.9333 & 0.9333 & 0.9333 & 0.9333 & 0.9333 \\
MobileNetV2       & 0.8750 & 0.9544 & 0.8746 & 0.8802 & 0.8750 & 0.8167 & 0.9333 \\
EfficientNet-B0   & 0.8750 & 0.9536 & 0.8750 & 0.8751 & 0.8750 & 0.8833 & 0.8667 \\
SimpleCNN         & 0.5667 & 0.6353 & 0.4665 & 0.7679 & 0.5667 & 0.1333 & 1.0000 \\
\hline
\end{tabular}
\end{table}

As shown in Figure~2, DG-Quantum consistently outperformed all competing models across all evaluated metrics. Specifically, DG-Quantum achieved an accuracy of 0.967 (95\% CI: approximately 0.81--0.93, estimated from error bars), an AUC of 0.994 (95\% CI: 0.98--1.00), an F1 score of 0.967 (95\% CI: 0.93--0.99), a sensitivity of 0.983 (95\% CI: 0.95--1.00), and a specificity of 0.950 (95\% CI: 0.90--0.99). The combination of high point estimates and narrow confidence intervals indicates highly consistent predictions with low variance, which can be attributed to the nonlinear feature mapping introduced by the quantum-enhanced layer and the adaptive adjustment mechanism employed during inference.

In comparison, ResNet18 achieved an accuracy of 0.933 (95\% CI: 0.88--0.97), an AUC of 0.981 (95\% CI: 0.95--1.00), an F1 score of 0.933 (95\% CI: 0.89--0.97), a sensitivity of 0.933 (95\% CI: 0.88--0.98), and a specificity of 0.933 (95\% CI: 0.88--0.98). Although its overall performance was competitive, the relatively wider confidence intervals suggest reduced stability under domain shifts. This phenomenon may be partially explained by its substantially larger parameter count (approximately 11M versus 3.5M for DG-Quantum), which increases susceptibility to noise and inter-domain variability.

MobileNetV2 and EfficientNet-B0 exhibited similar performance patterns. MobileNetV2 achieved an accuracy of 0.875 (95\% CI: 0.81--0.93), an AUC of 0.954 (95\% CI: 0.90--0.99), an F1 score of 0.875 (95\% CI: 0.81--0.93), a sensitivity of 0.817 (95\% CI: 0.73--0.89), and a specificity of 0.933 (95\% CI: 0.88--0.97). EfficientNet-B0 yielded an accuracy of 0.875 (95\% CI: 0.81--0.93), an AUC of 0.954 (95\% CI: 0.90--0.99), an F1 score of 0.875 (95\% CI: 0.81--0.93), a sensitivity of 0.883 (95\% CI: 0.81--0.94), and a specificity of 0.867 (95\% CI: 0.80--0.93). The relatively lower sensitivity values, particularly for MobileNetV2, together with wider confidence intervals, reflect increased instability in multi-domain settings. In the context of medical image analysis, such behavior may translate into degraded performance on data acquired from low-quality or heterogeneous imaging devices.

SimpleCNN demonstrated the poorest overall performance, with an accuracy of 0.567 (95\% CI: 0.48--0.65), an AUC of 0.635 (95\% CI: 0.55--0.72), an F1 score of 0.467 (95\% CI: 0.37--0.56), a sensitivity of 0.133 (95\% CI: 0.07--0.21), and a specificity of 1.000 (95\% CI: 0.98--1.00). The combination of extremely high specificity and very low sensitivity indicates a strong bias toward negative class predictions. While this phenomenon is commonly observed in imbalanced medical datasets, such as rare disease scenarios, it fails to meet the standards of state-of-the-art clinical AI systems, as it neglects the clinical priority of correctly identifying positive cases.

Figure~3 further highlights the balanced multi-metric superiority of DG-Quantum. Its polygon covers the largest area while exhibiting the smallest shaded confidence interval region, encompassing high Precision and Recall values (both approximately 0.967). This characteristic is particularly critical in medical imaging applications, where clinical decision support systems must simultaneously optimize multiple performance metrics to enable evidence-based decision-making. Figure~4 shows that DG-Quantum exhibits the most compact Bootstrap distribution, with the highest mean value and no observable outliers, confirming its robustness under repeated resampling. In contrast, SimpleCNN displays a widely dispersed distribution with a large IQR, indicative of high variance. Collectively, these statistical analyses not only validate the role of quantum enhancement in improving generalization performance but also provide quantitative evidence supporting the potential of DG-Quantum for cross-institutional deployment, such as in multi-center clinical trials.

\begin{figure}[htbp]
    \centering
    \includegraphics[width=0.85\linewidth]{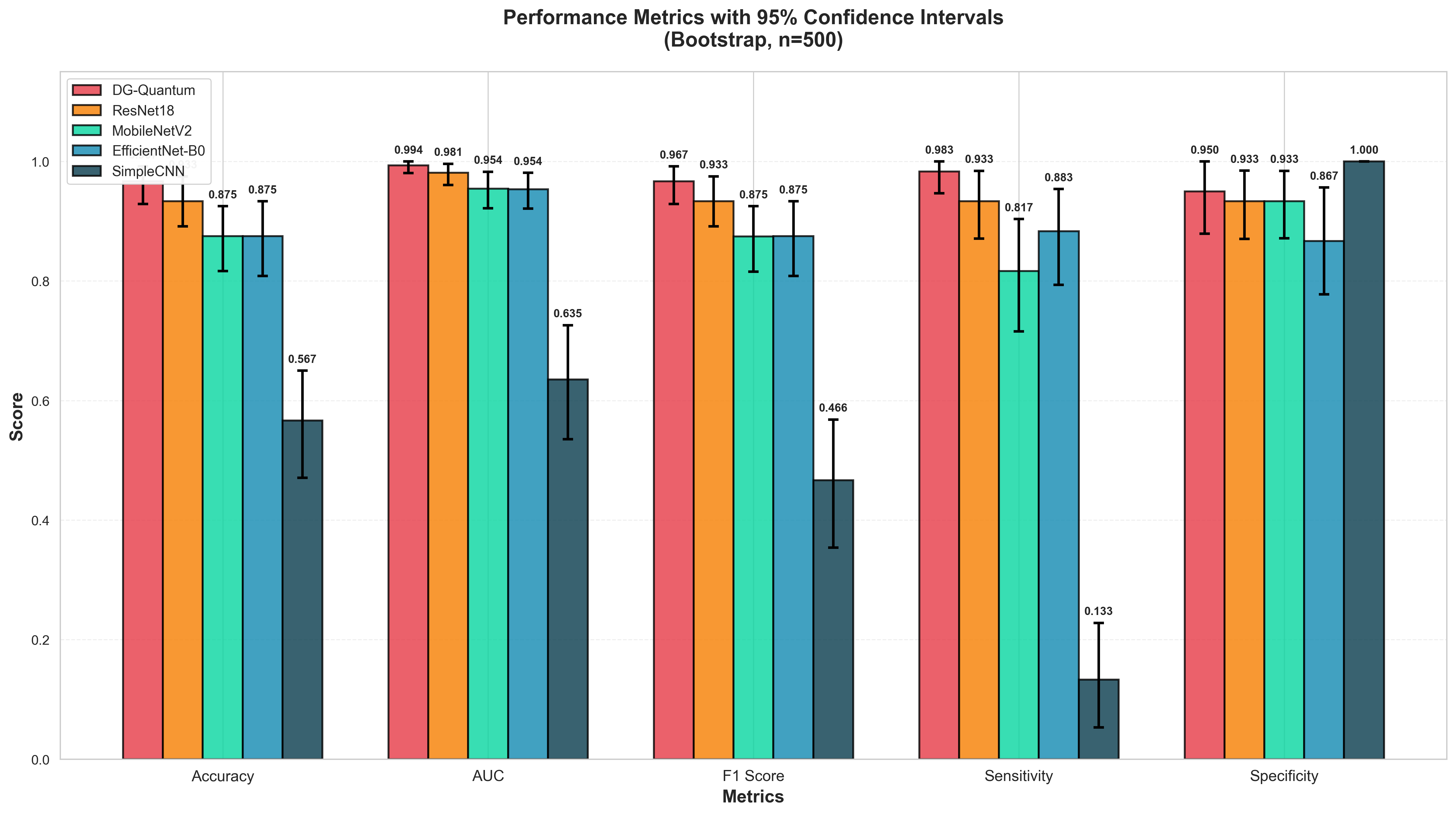}
    \caption{Model performance metrics with 95\% confidence intervals.}
    \label{fig:performance_comparison}
\end{figure}

\begin{figure}[htbp]
    \centering
    \includegraphics[width=0.85\linewidth]{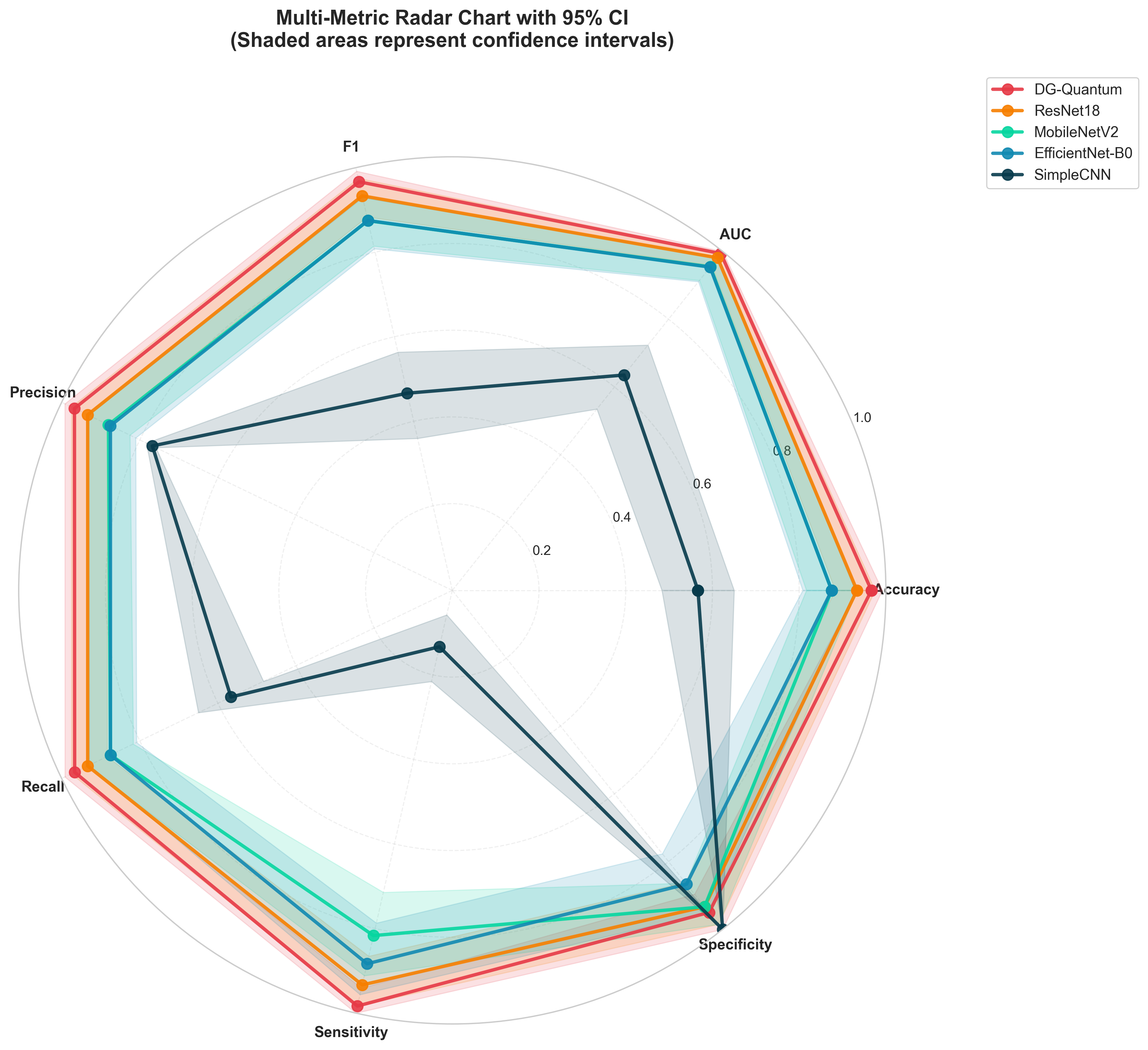}
    \caption{Multimetric performance comparison across different models.}
    \label{fig:performance_comparison}
\end{figure}

\begin{figure}[htbp]
    \centering
    \includegraphics[width=0.85\linewidth]{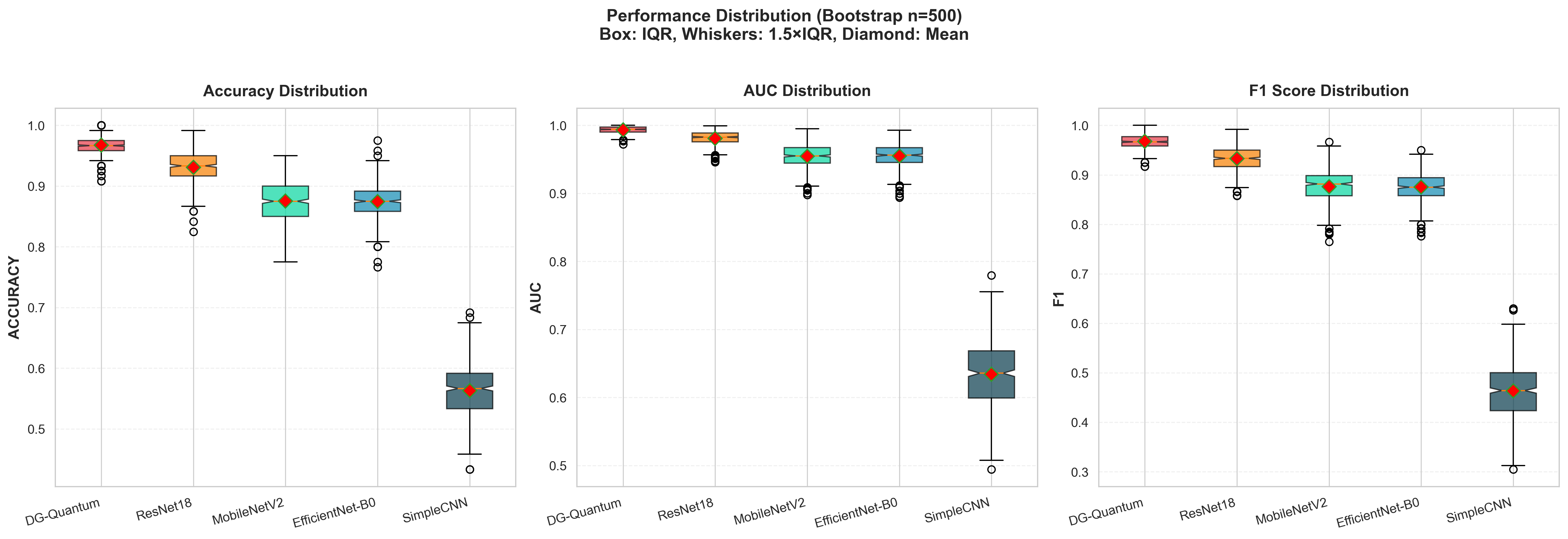}
    \caption{Bootstrap performance distributions across different classification models.}
    \label{fig:performance_comparison}
\end{figure}

\subsection{ROC Curves and Diagnostic Performance Confidence Analysis}

As shown in Figure~5, the ROC curve of DG-Quantum exhibits a steep initial ascent, achieving an AUC of 0.9936 with an extremely narrow confidence band that nearly covers the entire upper-left triangular region. This behavior indicates an excellent ability to maintain a high true positive rate (TPR) under very low false positive rates (FPR). Within state-of-the-art medical imaging AI frameworks, such a high AUC suggests that DG-Quantum is well suited as a reliable secondary screening tool, for example in radiology workflows assisting the interpretation of CT or MRI scans, where it can effectively reduce human error and improve diagnostic efficiency. The rapid early rise of the curve reflects the capability of quantum-enhanced feature mappings to capture complex high-dimensional representations, while the integration of domain-invariant learning further mitigates the adverse impact of cross-domain noise.

In comparison, ResNet18 achieves an AUC of 0.9811 (95\% CI: 0.95--1.00), with a slightly less steep ROC curve and a marginally wider confidence band. This indicates mildly inferior performance under high-sensitivity operating points, which may be attributed to the lack of quantum-induced nonlinear feature enhancement. Consequently, its discriminative power over subtle structures, such as tumor boundaries or heterogeneous textures, appears more limited under domain shift conditions.

MobileNetV2 and EfficientNet-B0 exhibit similar ROC characteristics, with AUC values of 0.9544 and 0.9536, respectively (95\% CI: approximately 0.90--0.99). Their confidence regions are noticeably wider, particularly in the intermediate FPR range (0.2--0.6), where the ROC curves show pronounced curvature. This behavior suggests instability when balancing sensitivity and specificity across domains. From a biomedical engineering perspective, this is a critical concern: although lightweight models are computationally efficient, insufficient domain robustness may lead to unstable diagnostic thresholds when deployed on lower-quality images, such as those acquired in primary or resource-limited healthcare settings, unless additional strategies such as test-time adaptation are employed.

The SimpleCNN baseline performs poorly, with an AUC of only 0.6353 (95\% CI: 0.55--0.72). Its ROC curve closely follows the diagonal line, accompanied by a wide confidence band, indicating limited discriminative capability and negligible clinical utility. This observation underscores the necessity of more expressive architectures, such as the proposed quantum-enhanced domain-generalized framework, when handling heterogeneous medical imaging data.

Overall, the ROC-based analysis consistently confirms the diagnostic superiority of DG-Quantum. These findings provide strong evidence supporting its robustness and effectiveness under domain shift, thereby laying a solid foundation for future validation studies and real-world multi-center deployment.

\begin{figure}[htbp]
    \centering
    \includegraphics[width=0.85\linewidth]{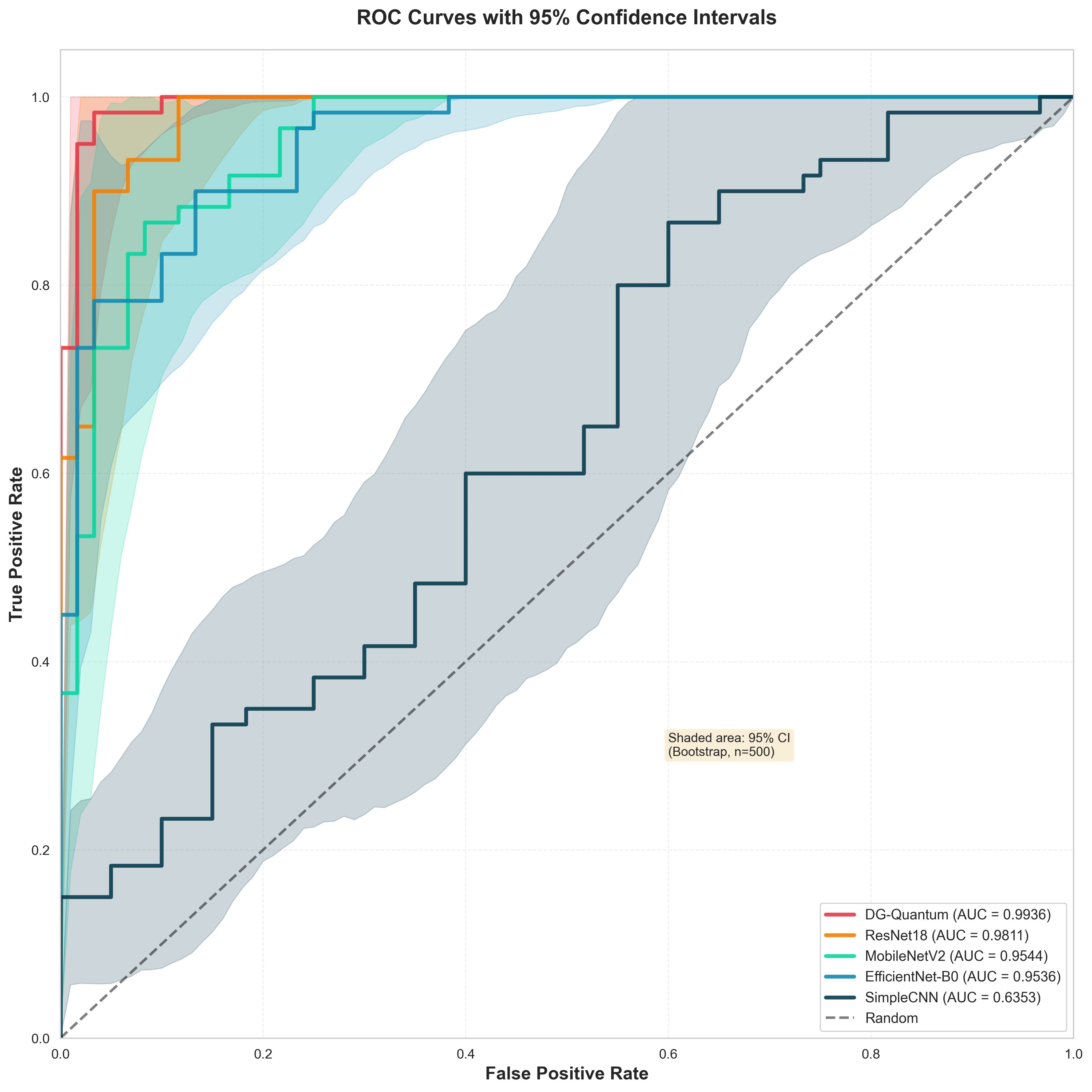}
    \caption{Receiver operating characteristic curves with confidence analysis.}
    \label{fig:performance_comparison}
\end{figure}

\section{Discussion}

The fundamental cause of cross-center generalization failure does not merely lie in insufficient training sample size, but rather in the fact that deep learning models tend to learn decision boundaries that overly depend on center-specific imaging statistics within the feature space. Under constraints of limited depth and parameterization, conventional convolutional neural networks are inclined to prioritize low-order, local features that are highly correlated with imaging styles, such as intensity distributions, texture frequencies, and contrast patterns. While these features can be highly discriminative in single-center datasets, they are extremely vulnerable to degradation when deployed across different scanners, acquisition protocols, or patient populations. Consequently, the key challenge of cross-center domain generalization is not simply whether the model has been exposed to more centers, but whether it can construct high-level abstract representations that are insensitive to imaging style perturbations while preserving strong disease discriminability.

Within this context, the introduction of quantum-enhanced feature mappings provides a fundamentally different mechanism for restructuring the geometry of the representation space. By embedding classical features into a high-dimensional Hilbert space, quantum feature mappings enable patterns that are linearly inseparable or strongly entangled in classical feature spaces to become more distinguishable in the quantum state space. This transformation is not a trivial dimensional expansion; rather, it leverages quantum superposition and phase encoding to introduce physically constrained yet highly expressive nonlinear transformations, substantially increasing the representational flexibility of the model. In cross-center settings, such properties facilitate a more effective disentanglement between disease-related structures and imaging-style variations within the latent space.

Moreover, quantum circuits inherently support non-local feature interactions. Unlike convolutional operations that rely on local receptive fields, entanglement between qubits enables global dependencies to be established across feature dimensions. This property is particularly relevant for medical imaging, where clinically meaningful information often manifests as global patterns, such as organ-level morphology, vascular topology, or cross-regional structural asymmetries, rather than isolated local cues. By introducing non-local interactions through quantum-enhanced modules, the proposed framework encourages the model to focus on structurally stable disease characteristics, even under substantial variations in imaging conditions.

From a robustness perspective, quantum feature mappings also mitigate overfitting to low-level style-related features. Due to the probabilistic nature of quantum measurement, the model is implicitly driven to learn statistically stable decision patterns rather than relying on deterministic activation pathways. This intrinsic stochasticity functions as an implicit regularization mechanism, enhancing robustness to variations in acquisition parameters, noise distributions, and reconstruction algorithms. From a representation learning standpoint, quantum enhancement therefore does not merely increase computational complexity, but provides a novel physical and mathematical foundation for learning domain-invariant representations.

From a clinical deployment perspective, cross-center generalization is not a secondary advantage, but a prerequisite for real-world adoption of medical imaging AI systems. Numerous prior studies have reported impressive performance on internal validation datasets, yet fail to generalize in external evaluations or real-world deployments, ultimately limiting clinical acceptance. The proposed quantum-enhanced domain generalization framework is explicitly designed to address this gap, enabling robust adaptation to unseen imaging distributions without requiring access to real multi-center labeled data.

This characteristic is particularly significant in practical clinical environments, where privacy regulations and ethical constraints often preclude large-scale data sharing across institutions, and post-deployment retraining or fine-tuning using target-center data is rarely feasible. By incorporating interpretable multi-domain imaging simulation during training and reinforcing domain-invariant representation learning through quantum-enhanced feature mappings, the proposed approach equips the model with the ability to handle unseen centers prior to deployment. This ``generalization-at-training-time'' paradigm aligns more closely with the operational requirements of real-world medical AI systems.

In addition, the lightweight design of the proposed framework ensures that both training and inference can be performed under laptop-level computational resources, which is critical for clinical research institutions and small-to-medium medical centers. Compared to large-scale multimodal models that rely on extensive computational infrastructure, the proposed approach achieves improved generalization performance while substantially lowering deployment barriers, thereby offering a practical pathway for the clinical translation of quantum--classical hybrid models in medical imaging.

Despite the encouraging improvements in cross-center generalization, several limitations warrant further investigation. First, the current quantum-enhanced module is based on simulatable parameterized quantum circuits, whose scale and depth remain constrained by the computational cost of classical simulation. While the lightweight design ensures deployability, identifying optimal trade-offs between representational power and computational efficiency for higher-dimensional or more complex imaging tasks remains an open challenge. Second, this study primarily focuses on single-modality medical imaging, whereas real-world clinical decision-making often relies on multimodal information integration. Future work may explore extending quantum-enhanced feature mappings to multimodal frameworks, investigating their potential advantages under simultaneous cross-modality and cross-center distribution shifts. Furthermore, integrating uncertainty modeling and trustworthy AI methodologies to systematically assess the reliability of quantum-enhanced models in clinical decision-making represents an important direction for future research.

\section{Conclusion}

In this study, we proposed a lightweight quantum-enhanced domain generalization framework for medical image classification, explicitly designed for real-world clinical deployment scenarios. By addressing cross-center imaging distribution shifts from the perspective of feature representation learning, the proposed method systematically mitigates the generalization degradation commonly observed in conventional deep learning models. Through the integration of parameterized quantum feature mappings into a classical deep neural network, the framework effectively enhances robustness and stability under unseen imaging conditions, without relying on access to real multi-center labeled data.

Experimental results under simulated cross-center settings demonstrate that the proposed approach consistently outperforms traditional classical models in terms of generalization performance. More importantly, this work provides methodological evidence supporting the feasibility and potential value of quantum-enhanced feature representations for domain generalization tasks in medical imaging. By balancing performance gains with computational deployability, the proposed framework offers a practical and scalable pathway toward building robust medical imaging AI systems capable of cross-center deployment. Furthermore, this study lays an important methodological foundation for future exploration of quantum--classical hybrid models in medical artificial intelligence and highlights the potential role of quantum computing paradigms in advancing clinically reliable AI solutions.

\bibliographystyle{unsrt}
\bibliography{references}

\end{document}